# MNIST-Fraction: Enhancing Math Education with AI-Driven Fraction Detection and Analysis

Pegah Ahadian
Kent State University
Kent, Ohio, USA
pahadian@kent.edu

Yunhe Feng
North Texas University
Denton, Texas, USA
yunhe.feng@unt.edu

Karl Kosko
Kent State University
Kent, Ohio, USA
kkosko1@kent.edu

Richard Ferdig
Kent State University
Kent, Ohio, USA
rferdig@kent.edu

Qiang Guan
Kent State University
Kent, Ohio, USA
qguan@kent.edu

## ABSTRACT

Mathematics education, a crucial and basic field, significantly influences students' learning in related subjects and their future careers. Utilizing artificial intelligence to interpret and comprehend math problems in education is not yet fully explored. This is due to the scarcity of quality datasets and the intricacies of processing handwritten information. In this paper, we present a novel contribution to the field of mathematics education through the development of MNIST-Fraction, a dataset inspired by the renowned MNIST, specifically tailored for the recognition and understanding of handwritten math fractions. Our approach is the utilization of deep learning, specifically Convolutional Neural Networks (CNNs), for the recognition and understanding of handwritten math fractions to effectively detect and analyze fractions, along with their numerators and denominators. This capability is pivotal in calculating the value of fractions, a fundamental aspect of math learning. The MNIST-Fraction dataset is designed to closely mimic real-world scenarios, providing a reliable and relevant resource for AI-driven educational tools. Furthermore, we conduct a comprehensive comparison of our dataset with the original MNIST dataset using various classifiers, demonstrating the effectiveness and versatility of MNIST-Fraction in both detection and classification tasks. This comparative analysis not only validates the practical utility of our dataset but also offers insights into its potential applications in math education. To foster collaboration and further research within the computational and educational communities. Our work aims to bridge the gap in high-quality educational resources for math learning, offering a valuable tool for both educators and researchers in the field.

## 1 INTRODUCTION

Introducing MNIST-Fraction, a breakthrough in individual dataset and deep learning, dedicated to transforming how math problems are detected and understood. Mathematics is the foundation of numerous scientific and engineering disciplines, and accurate math problem detection plays a crucial role in streamlining workflows and improving productivity. In the realm of math problem detection, traditional approaches have encountered notable challenges when it comes to handling a wide array of fonts, styles, and formats. These challenges often lead to errors and inaccuracies in the detection process. Prior efforts in this domain have strived to address these issues, but they have fallen short in achieving comprehensive and reliable solutions. Several existing works have attempted to tackle the challenges associated with diverse fonts, styles, and formats in math problem detection. These works typically employ rule-based methods, template matching, or simple feature-based techniques. While these approaches can work reasonably well on standardized and well-structured documents, they struggle when confronted with variations in handwriting, unconventional symbols, or non-standard mathematical notations. Rule-based methods, for instance, rely on predefined rules to identify specific patterns or symbols in the text. However, they lack the flexibility needed to handle the vast diversity of fonts and styles found in real-world documents. Feature-based methods leverage characteristics like the size, shape, and spatial distribution of symbols within the text. However, these techniques are sensitive to changes in formatting and may fail to generalize across a wide range of fonts and styles. Math problem detection is no trivial task. Unlike generic object detection, recognizing mathematical expressions requires an understanding of complex symbols, equations, and their structural relationships. Conventional computer vision methods may struggle with these intricacies, leading to suboptimal performance.

In light of these limitations, there is a clear need for an innovative approach that can overcome the challenges posed by diverse formats of handwritten in math problem detection. MNIST-Fraction is being presented as the pioneering solution that directly addresses these limitations by introducing a novel methodology. A fundamental aspect of MNIST-Fraction's success is the unique dataset we have compiled for training and evaluation purposes. This dataset encompasses an array of fraction math problems, sourced from various handwritten and styles of digits and fractions. It covers real-world examples, ensuring robustness and versatility. The dataset's exclusivity stems from the careful curation process, where our team of experts meticulously labeled the expression, in this version with focus on fraction, to create ground truth annotations. This attention to detail ensures that MNIST-Fraction is well-equipped to handle real-world scenarios, making it a valuable tool for researchers and educators [8, 22, 23]. MNIST-Fraction has undergone testing and evaluation, proving its mettle as a first benchmark datasets in math-fraction detection. Presented work, dataset and suggested model, consistently outperforms existing methods, achieving unparalleled accuracy and speed in math problem detection, Figure 4. Its applications are widespread, with significant potential in diverse fields:

**Automated Grading Systems:** MNIST-Fraction can revolutionize the educational landscape by powering automated grading systems [16]. With its ability to accurately evaluate students' mathematical responses, educators can save valuable time and provide more personalized feedback to learners.

**Document Analysis:** Incorporating MNIST-Fraction into document analysis [3] workflows streamlines the process of digitizing mathematical content. By automatically detecting math problems, researchers and archivists can efficiently process large volumes of documents, enabling faster access to critical information.

**Educational Technology:** Integrating MNIST-Fraction into educational technology [11] platforms empowers learners with real-time feedback on their math exercises and assignments. This interactive learning experience enhances comprehension and encourages active participation.

MNIST-Fraction represents a pioneering leap in the field of computer vision and deep learning. With its exclusive dataset, the suggested method delivers unmatched accuracy and efficiency in math problem detection specifically fractions which are a fundamental concept for students to later succeed in various algebraic topics in high school and college. "fractions are perceived as one of the most difficult areas in school mathematics to learn and teach" [6, 9]. One area of significant potential in supporting teachers in the classroom is to provide them with tools to provide specific feedback related to children's level of reasoning with fractions. From automating grading systems to enhancing document analysis and educational technology, MNIST-Fraction has the potential to transform various industries.

## 2 RELATED WORK

Math recognition has witnessed substantial advancements in recent years, driven by the exploration of novel techniques and the emergence of benchmark competitions [7]. Early efforts in handwritten math recognition were largely based on rule-based systems that relied on explicit knowledge of symbol shapes and spatial relationships [2]. However, these methods struggled to handle the inherent variability in handwriting styles and symbol representations. The integration of deep learning techniques and the focus on fractions within MNIST-Fraction contribute to the evolving landscape, while drawing insights from related research such as Richard Zanibbi work [19]. This paper discusses the wider spectrum of mathematical expression recognition, encompassing various notations, symbols. The advancements explored in this paper, offer valuable insights that can complement the efforts of MNIST-Fraction.

In handwritten mathematical expression recognition, X Zhang [20] has introduced an innovative perspective by advocating for a coarse-grained recognition approach. By categorizing mathematical symbols into broader groups, this work demonstrates enhanced generalization capabilities, making the recognition task more efficient and adaptable and [13]. This notion complements the goals of "MNIST-Fraction" by fostering a deeper comprehension of fraction components, further refining the interpretation of handwritten fractions. Ye Yuan Yuan et al. [18] presented the Syntax-Aware Network for Handwritten Mathematical Expression Recognition. This work contributes to the advancement of mathematical expression recognition by leveraging syntax-aware techniques within neural network architectures. Also Bohan Li [12] by incorporating counting-aware techniques into the network architecture aims to improve the accuracy and efficiency of recognizing and understanding handwritten mathematical expressions, with a particular focus on the counting aspect. It is important, how to use the different technologies along with the methods, like "Segmentation and Contour Detection for handwritten mathematical expressions using OpenCV" [10] that introduced a method for segmenting and detecting contours in handwritten mathematical expressions through the utilization of the OpenCV library. The approach is to process handwritten expressions by breaking them down into distinct components, aided by OpenCV's tools for segmentation and contour detection.

Nowadays, widely used neural language models commonly employ BERT [14, 21] or RNN for modeling purposes. In the case of mathematical expression language, its linguistic characteristics encompass not only the connections between mathematical symbols but also encompass the behavior within symbols and the structural relationships they hold. By employing deformable convolutions and hybrid architectures, DeHyFoNet [1] aims to improve the precision of formula detection in scanned document images. Identifying mathematical content present in documents by utilizing the U-Net architecture and training it on a varied dataset to enhance the precision of mathematical expression detection [15]. Xuedong Tian [17] introduce a formula embedding model based on BERT to streamline formula retrieval in ARQMath2 tasks. To demonstrate the model's efficacy for mathematical language processing. This model processes LaTeX formulas as input and generates fixed-dimensional embeddings.

## 3 MNIST-FRACTION DATASET

In order to train a deep learning model that can detect fractions efficiently and effectively, we first construct a fraction dataset mainly using the publicly available MNIST datasets [4]. As the MNIST dataset consists of thousands of handwritten digits ranging from 0 to 9, we believe it is a good data source to build our fraction dataset. The purpose of this task is creating a dataset of synthetic fraction images for research and educational purpose, likely involving machine learning and computer vision tasks. The images are generated by combining individual digit images to form fraction representations. These synthetic datasets can be useful for training and testing machine learning models related to image recognition, classification, or understanding fractions.

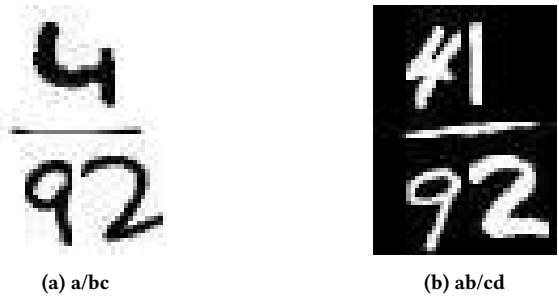

(a) a/bc  (b) ab/cd

Figure 1: Generated Dataset Samples of Inverted Single/ Double Digit Fraction a/bc and Double Digits Fraction ab/cd

## 3.1 Preprocess Digits in MNIST

In this section, we present a algorithm to generate synthetic fraction representations using images sourced from the MNIST dataset. The objective is to create fraction images in the format "a/b" where "a" represents a single-digit numerator and "b" represents a single-digit denominator, and double digits fraction "a/bc", "ab/cd" Figure 1b. By utilizing the rich set of digit images available in the MNIST dataset, we can create diverse and customizable fraction representations for various educational or research purposes. The presented algorithm consists of two main functions: generate-fraction and run-multiple-times. This algorithm offers flexibility in generating synthetic fraction images tailored to specific needs. By selecting different digit images for the numerator and denominator, along with diverse fraction bar images, a wide range of fraction representations can be generated. These synthetic fractions can be valuable for educational materials, visualizations, and research.

## 3.2 Fraction Bar Design

In this section, we present a method to generate a fraction bar based on image processing techniques and using the digit "1" from the MNIST dataset. A fraction bar is a crucial visual element in representing fractions. The algorithm's unique approach involves repurposing the digit "1" from the MNIST dataset as a horizontal fraction bar. By leveraging the inherent shape of the digit "1" which resembles a horizontal line, we create a recognizable and consistent fraction bar across various contexts. This method offers a visually appealing and recognizable alternative to traditional fraction bar representations. To ensure optimal alignment and aesthetic appeal, the fraction bar image undergoes a two-fold transformation. First, it is resized to span the calculated width, fostering harmonious integration with the other elements. Second, the fraction bar is transposed, making it horizontally aligned, thus guaranteeing a seamless fit within the fraction image. Incorporating the digit "1" from the MNIST dataset into Fraction Bar Design offers a noteworthy advantage by facilitating the creation of fraction bars that closely resemble those found in real-world handwritten problems.

## 3.3 Construct Fractions

The process of constructing fractions involves the innovative utilization of digital imagery to visually represent numerical relationships. Leveraging the power of image processing techniques, the generate-fraction-array and run-multiple-times functions contribute to the creation of compelling fraction representations. The point of presented dataset is the method of generation. We define a function which takes as an input paths to digit images representing the numerator "a" the two parts of the denominator "b", the fraction bar, and the spacing between the "a" and "b" parts of the denominator. It produces a 2D array (image representation) of the fraction a/b. The generate-fraction-array function takes advantage of the flexibility provided by images. By employing three input images – the numerator, denominator, and fraction bar – this function dynamically generates fraction representations in the format "a/b". The underlying algorithm orchestrates the precise alignment of these images, ensuring a coherent and visually pleasing result. By transposing the fraction bar and strategically positioning the images, the function constructs fraction images that vividly convey numerical proportions. Furthermore, the run-multiple-times function amplifies the impact of this approach by enabling the generation of diverse samples. With parameters such as the numerator, denominator, and the number of samples to generate, this function fosters the creation of a range of fraction representations. Each sample encapsulates unique combinations of images, enriching the dataset with visual diversity, Algorithm 1.

The extra process on generated fraction which we present is color inversion, see Figure 1b. This color inversion step simplifies the digit recognition task for machine learning models, as black digits on a white background, see Figure 1a, provide more distinguishable features for the algorithms to learn from for more compatibility with student hand written.

## 4 EXPERIMENTAL EVALUATION

In this section, we provide some experimental results in order to classification tasks, we present Table 1 of different classifiers to form a benchmark on MNIST-Fraction dataset. As MNIST is a well-established benchmark in the field of machine learning for handwritten digit recognition, comparing MNIST-Fraction to MNIST allows researchers to gauge the performance, effectiveness, and robustness of the MNIST-Fraction dataset in a familiar context. Also we peresent a CNN model which trained on 72159 images of MNIST-Fraction to show the performance of model on this dataset in math fraction detection and fraction's details. The comparison offers insights into potential applications of the MNIST-Fraction dataset in math education and validates its practical utility. By showing that MNIST-Fraction can handle more complex scenarios than MNIST, we encourage its adoption for educational purposes, promoting a better understanding of handwritten fractions. To further improve the model's generalization and robustness to variations in handwritten fractions, we employed data augmentation techniques using

---

**Algorithm 1:** Fraction Generation $\frac{a}{b}$

**Require:** $a, b \in \mathbb{N}$ where $b \neq 0$
**Ensure:** A dataset of image-represented fractions $\frac{a}{b}$
  **Initialization:**
  Define digit set $\{D \in \mathbb{Z} \mid 1 \leq D \leq 9\}$ for both numerators and denominators.
  **Generate Fraction Images:**
  **for** $a_i, b_j$ in $D \times (D \setminus \{0\})$ **do**
    Generate fraction image $I_{\frac{a_i}{b_j}}$ from digit images.
    Apply color inversion.
    Store the processed image.
  **end for**
  **Create Dataset:**
  **for** $k = 1$ to $n$ **do**
    Select random $a_i, b_j$ from $D$.
    Position and merge $I_{\frac{a_i}{b_j}}$ into a target image array $F$.
    Store $F$ with varied alignments and spacing for diversity.
  **end for**

---

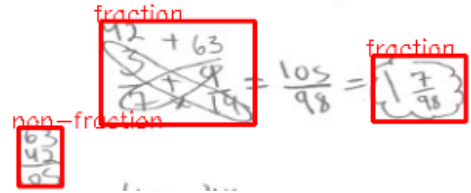

**Figure 2:** Student Handwritten Fraction Detection by CNN, Trained on MNIST-Fraction Dataset

ImageDataGenerator that includes, Zoom, Horizontal shift(up to 5 percent), and Vertical shift. We are actively exploring additional data augmentation techniques like rotations, flips, and noise injection to potentially further improve performance. The dataset was split into train/test/validation sets using stratified random sampling with a 70/15/15 split. This ensures that the class proportions (different fraction structures) are preserved in each set. Additionally, the data was shuffled before splitting and a random seed was set to guarantee fairness and reproducibility across evaluations.

Table 1 provides a detailed comparison of various classifiers and their respective parameter configurations, tested on two datasets: MNIST-Fraction and MNIST. The MNIST-Fraction contains 11 classes, 10 individual digits (0-9) for potential numerators and denominators, and 1 additional class for identifying the presence and structure of the fraction (single/multiple numerator, single/multiple denominator).

The table presents a wide range of classifiers, including Decision Trees, Extra Trees, Random Forest, SVC, GaussianNB, KNeighbors, Perceptron, Passive Aggressive, SGDClassifier, and Logistic Regression. Each classifier is tested with different parameters like criterion (crit.), max depth (m-d.), split strategy (splt.), kernel types, penalty types, and others to optimize performance.

For each classifier and parameter setting, the table reports performance on the MNIST-Fraction and MNIST datasets. Performance is measured in terms of classification accuracy. The classifiers exhibit varied performance, with some configurations achieving high accuracy, indicating effective parameter tuning and classifier choice for the specific characteristics of the MNIST datasets. The detailed breakdown allows for a nuanced understanding of how each classifier and its parameters perform, providing insights into the most effective models and settings for handwriting recognition tasks like those presented in MNIST. This comprehensive analysis helps in identifying the most promising approaches for further tuning and application in similar tasks.

The comparison offers insights into potential applications of the MNIST-Fraction dataset in math education and validates its practical utility. By showing that MNIST-Fraction can handle more complex scenarios than MNIST, the authors can encourage its adoption for educational purposes, promoting a better understanding of handwritten fractions. All algorithms are repeated 5 times by shuffling the training data and the average accuracy on the test set is reported. The benchmark on the MNIST dataset is also included for a side-by-side comparison.

Also, we present our approach to detecting fractions in student answer sheets using Convolutional Neural Networks and the MNIST-Fraction dataset. The successful application of CNNs in computer vision tasks, combined with the utilization of a specialized fraction dataset, has yielded promising results in automating the process of fraction detection. CNNs [5] have demonstrated remarkable capabilities in analyzing and processing visual data. These neural networks are particularly suited for tasks involving image recognition, segmentation, and classification due to their ability to learn hierarchical features directly from the raw image data. This hierarchical feature extraction enables CNNs to capture intricate patterns and relationships within images, making them an ideal choice for our task of fraction detection. For our fraction detection task, we leveraged the MNIST-Fraction dataset, a specialized collection of labeled images containing various instances of fractions. This dataset was meticulously curated to encompass a diverse range of fraction representations found in student answer sheets. It includes fractions written in different handwriting styles based on MNIST. To perform fraction detection, we first preprocessed the MNIST-Fraction dataset, ensuring that the images were appropriately scaled and centered for optimal training. We then designed and trained a CNN architecture tailored for fraction detection. The model is a Convolutional Neural Network designed for the recognition of handwritten math symbols and digits, utilizing a Sequential architecture in TensorFlow's Keras API. It comprises an input layer accepting 56x56 grayscale images, three convolutional layers each followed by ReLU activation and MaxPooling for feature extraction and dimensionality reduction, and a Flatten layer to transform 2D features into a 1D vector. To prevent overfitting, a Dropout layer is included before three fully connected (Dense) layers that culminate in a softmax output for binary classification. This architecture employs L2 regularization and Glorot uniform initialization to ensure stable and effective learning, representing a balanced approach to capturing complex patterns in the data while maintaining model generalizability and efficiency. The CNN was trained on a subset of the MNIST-Fraction dataset, Figure 2 with careful consideration given to data augmentation techniques to improve the model's generalization. Upon completion of the training phase, we evaluated the CNN on a separate test set of MNIST-Fraction images as well as real-world student answer sheets. The results were highly promising, with the CNN demonstrating a strong ability to accurately locate and delineate fractions within the answer sheets. Our model's performance was further validated by comparing its predictions with manually annotated ground truth. The results indicate an exemplary performance of a classification model across 11 classes(0 to 9 digits and fraction), with precision, recall, and F1-scores predominantly at or near perfection (0.99 or 1.00), signifying accurate predictions and comprehensive coverage of relevant instances. Overall, with an accuracy of 0.99 and similarly high macro, 0.99, and weighted averages, 0.99 for the F1-score, the model demonstrates outstanding consistency and reliability in its classification capabilities across a diverse set of categories. We showcase qualitative examples of MNIST-Fraction on our CNN model performance, with precision, recall, and F1-scores predominantly near perfection 0.99 illustrating successful fraction detection, accurate understanding of components, digit detection within fractions, and precise fraction value computation. These examples provide visual evidence of the model's efficacy in handling diverse fraction scenarios, see Figure 3 and Figure 4 and the model demonstrates exceptional accuracy, consistently achieving near-perfect precision, recall, and F1 scores across various classes, underscoring its effectiveness and precision in classification.

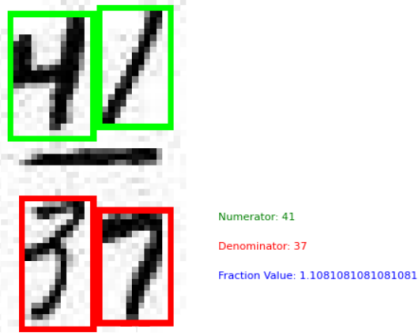

Figure 3: Numerator and Denominator Value Detection by Convolutional Neural Networks (CNNs)

## 5 CONCLUSION

Looking towards the future, the evolution of the MNIST-Fraction dataset and mathematics problem understanding holds exciting prospects for advancing various areas of deep learning. As part of our ongoing research, we are committed to presenting an updated iteration of the MNIST-Fraction dataset, enriched with additional features and complexities. This updated dataset will be accompanied by a thorough evaluation using diverse pre-trained models in the domain of deep learning, extending beyond math problem detection to encompass object detection, computer vision tasks, and machine learning tasks such as classification. By subjecting the dataset to a broader spectrum of tasks, we aim to demonstrate its versatility and applicability across multiple domains, showcasing its potential impact on a wider range of AI applications. Moreover, our future work involves expanding the MNIST-Fraction dataset to include an even broader array of categories, fostering its adaptability to an even greater diversity of mathematical content and notation styles. Through these efforts, we envision the MNIST-Fraction project contributing to the advancement of cutting-edge research and applications in the fields of AI and mathematics alike. In conclusion, the use of deep learning and computer vision in detecting mathematical problems holds great promise for the field of education. By leveraging the power of artificial intelligence and visual perception, educators and researchers can enhance students' problem-solving skills and promote a deeper understanding of mathematical concepts. The newly generated dataset, MNIST-Fraction, combined with advanced deep learning models, opens up exciting possibilities for improving educational platforms and advancing mathematical research.

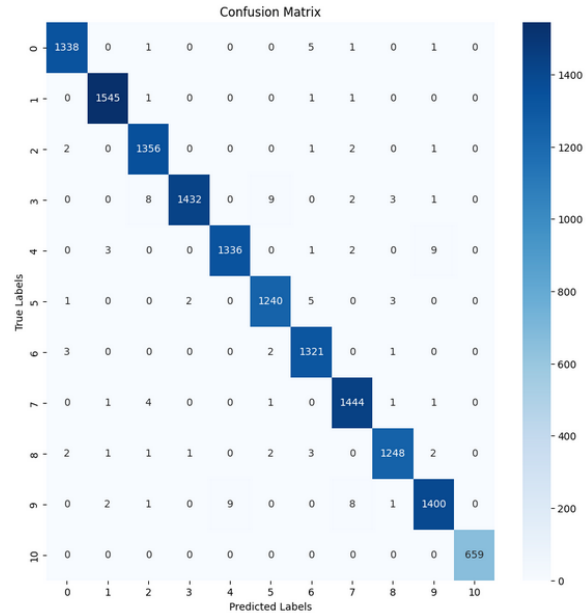

Figure 4: Confusion Matrix

Table 1: Benchmark on MNIST-Fraction and MNIST

| Classifier | Parameter | MNIST-Fraction | MNIST |
|---|---|---|---|
| DecisionTree | crit.=entropy, m-d.=10, splt.=best | 0.859 | 0.873 |
|  | crit.=entropy, m-d.=10, splt.=random | 0.843 | 0.861 |
|  | crit.=entropy, m-d.=50, splt.=best | 0.876 | 0.886 |
|  | crit.=entropy, m-d.=50, splt.=random | 0.860 | 0.877 |
|  | crit.=gini, m-d.=10, splt.=best | 0.827 | 0.853 |
|  | crit.=gini, m-d.=50, splt.=best | 0.861 | 0.873 |
|  | crit.=gini, m-d.=10, splt.=random |  |  |
|  | crit.=gini, m-d.=50, splt.=random |  |  |
| ExtraTrees | crit.=gini, m-d.=10 | 0.930 | 0.806 |
|  | crit.=gini, m-d.=50 | 0.970 | 0.845 |
|  | crit.=entropy, m-d.=10 | 0.935 | 0.810 |
|  | crit.=entropy, m-d.=50 | 0.970 | 0.847 |
| RandomForest | crit.=entropy, m-d.=10 | 0.948 | 0.950 |
|  | crit.=entropy, m-d.=50 | 0.967 | 0.969 |
|  | crit.=gini, m-d.=10 | 0.938 | 0.949 |
|  | crit.=gini, m-d.=50 | 0.967 | 0.968 |
| SVC | C=10, kernel=linear, gamma=scale | 0.931 | 0.927 |
|  | C=1, kernel=linear, gamma=scale | 0.943 | 0.929 |
|  | C=10, kernel=rbf, gamma=scale | 0.985 | 0.973 |
|  | C=1, kernel=rbf, gamma=scale | 0.978 | 0.966 |
|  | C=10, kernel=rbf, gamma=auto | 0.963 | 0.973 |
|  | C=1, kernel=rbf, gamma=auto | 0.944 | 0.966 |
|  | C=10, kernel=poly, gamma=scale | 0.980 | 0.976 |
|  | C=1, kernel=poly, gamma=scale | 0.977 | 0.957 |
|  | C=10, kernel=sigmoid, gamma=scale | 0.649 | 0.873 |
| GaussianNB | priors = 1 / 11 | 0.641 | 0.524 |
| KNeighbors | wt.=uniform, n-neighbors=5, p=1 | 0.967 | 0.957 |
|  | wt.=uniform, n-neighbors=9, p=2 | 0.970 | 0.943 |
|  | wt.=distance, n-neighbors=5, p=1 | 0.968 | 0.959 |
|  | wt.=distance, n-neighbors=9, p=2 | 0.971 | 0.944 |
| Perceptron | penalty=l1 | 0.850 | 0.887 |
|  | penalty=l2 | 0.837 | 0.845 |
|  | penalty=elasticnet | 0.836 | 0.845 |
| PassiveAggr. | C=1 | 0.867 | 0.877 |
|  | C=10 | 0.867 | 0.875 |
|  | C=100 | 0.867 | 0.880 |
| SGDClassifier | loss=hinge, penalty=l2 | 0.922 | 0.914 |
|  | loss=perceptron, penalty=l1 | 0.886 | 0.912 |
|  | loss=modified-huber, penalty=l1 | 0.899 | 0.910 |
|  | loss=modified-huber, penalty=l2 | 0.917 | 0.913 |
|  | loss=log-loss, penalty=elasticnet | 0.921 | 0.912 |
|  | loss=hinge, penalty=elasticnet | 0.915 | 0.913 |
| LogisticReg. | C=1, penalty=l2 | 0.929 | 0.917 |
|  | C=10, penalty=l2 | 0.928 | 0.916 |
|  | C=1, penalty=l1 | 0.929 | 0.917 |
|  | C=10, penalty=l1 | 0.928 | 0.909 |

## ACKNOWLEDGMENTS

Research reported here received support from the ASSISTments Foundation, Grant #1908159.